%% file: root.tex
\documentclass[letterpaper, 10 pt, conference]{ieeeconf}  

\IEEEoverridecommandlockouts                              

\usepackage{hyperref}
\hypersetup{bookmarksopen,bookmarksnumbered,
pdfpagemode=UseOutlines,
colorlinks=true,
linkcolor=blue,
anchorcolor=blue,
citecolor=blue,
filecolor=blue,
menucolor=blue,
urlcolor=blue
}
\usepackage{upgreek}
\usepackage{amsfonts,amssymb}
\usepackage{bm}
\usepackage{bbm}

\usepackage{amsthm} 
\usepackage{thmtools}
\usepackage{mathtools}
\usepackage{xspace}
\usepackage{units}
\usepackage{booktabs} 
\usepackage{tabulary}
\usepackage[usenames,dvipsnames,table]{xcolor} 
\usepackage{diagbox}
\usepackage[ruled,vlined,linesnumbered]{algorithm2e}  
\usepackage{parskip}
\SetKwComment{Comment}{$\triangleright$\ }{}
\usepackage{ifthen,version}
\usepackage{soul}
\usepackage{csquotes}
\usepackage{microtype}      
\usepackage{lipsum}
\usepackage{tikz}
\let\labelindent\relax
\usepackage{todonotes}
\usepackage{enumitem}
\usepackage[noadjust]{cite}
\usepackage{wrapfig}
\usepackage{graphicx}
\usepackage{array}
\usepackage{caption}
\usepackage{subcaption}

\usepackage{wrapfig}

\usepackage[normalem]{ulem}
\newcommand\redsout{\bgroup\markoverwith{\textcolor{red}{\rule[0.5ex]{2pt}{0.7pt}}}\ULon}

\newcommand{\xxnote}[3]{}
\ifx\hidenotes\undefined
  \usepackage{color}
  \renewcommand{\xxnote}[3]{\color{#2}{#1: #3}}
\fi

\usepackage{multirow}
\usepackage{pbox}

\newtheoremstyle{hypstyle}
{3pt} 
{3pt} 
{\itshape} 
{} 
{\bfseries} 
{.} 
{.5em} 
{} 

\theoremstyle{hypstyle}

\input{macros/math_operators}

\input{macros/math_defns}

\input{macros/text_shortcuts}
\graphicspath{{figs/}}

\title{\LARGE \bf
Do You See What I See? \\
Coordinating Multiple Aerial Cameras for Robot Cinematography
\vspace{-2mm}}

\author{Arthur Bucker$^{*1}$, Rogerio Bonatti$^{2}$, Sebastian Scherer$^{2}$ 
\vspace{-1.36mm}
\thanks{$^{1}$University of S\~ao Paulo, Brazil
        {\tt\small arthur.bucker@usp.br}}%
\thanks{$^{1}$The Robotics Institute, Carnegie Mellon University, Pittsburgh PA
        {\tt\small \{rbonatti, basti\}@cs.cmu.edu}}%
\thanks{$^{*}$ Supported by the CMU Robotics Institute Summer Scholars program}
}

\let\oldtwocolumn\twocolumn
\renewcommand\twocolumn[1][]{%
    \oldtwocolumn[{#1}{
    \begin{center}
           \includegraphics[width=1.0\textwidth]{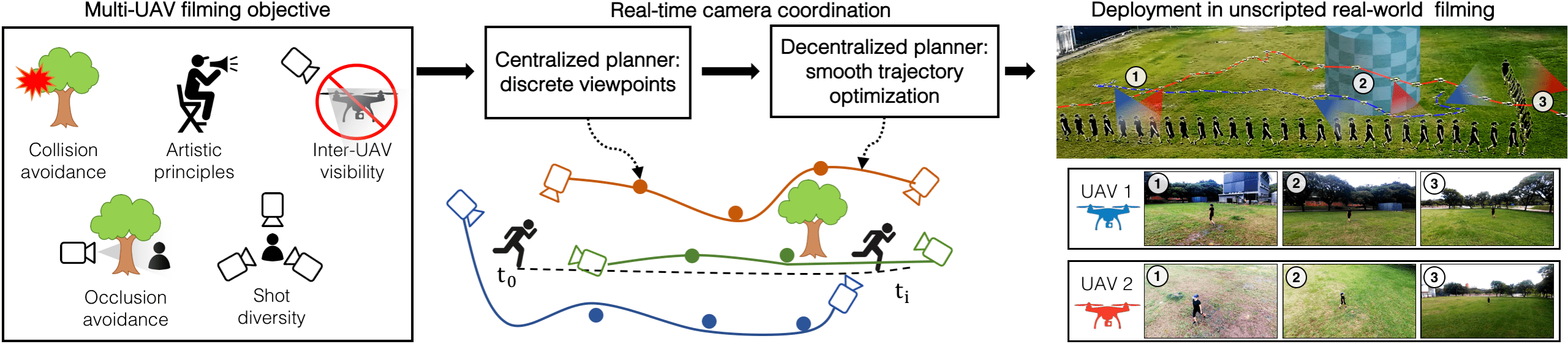}
           \captionof{figure}{\small Our framework uses a set of objectives for multi-UAV cinematography to coordinate all cameras trajectories with a fast centralized viewpoint planner coupled with a decentralized trajectory optimization algorithm. We deploy the system in real-world settings.
    	   \vspace{3.0mm}
    	   }
           \label{fig:main}
        \end{center}
    }]
}

\begin{document}

\maketitle
\thispagestyle{empty}
\pagestyle{empty}


\input{inputs/0_abstract}
\input{inputs/1_intro}
\input{inputs/2_related_work}
\input{inputs/3_problem_formulation}

\input{inputs/3_approach}
\input{inputs/4_exp}

\input{inputs/5_discussion}
\input{inputs/6_conclusion}

\section*{ACKNOWLEDGMENTS}

We thank CITI-Poli, Grupo Turing and the Skyrats team for the help with robot experiments. RB also gratefully acknowledges the support from the Waibel Education Fund.


\pagebreak

\footnotesize{
\bibliographystyle{IEEEtran}
\bibliography{IEEEexample}
}


\end{document}

%% file: macros/math_operators.tex
\DeclareMathOperator*{\argmin}{arg\,min}

\newcommand{\argminprob}[1]{\underset{#1}{\argmin}}

\newcommand{\real}[0]{\mathbb{R}}

\newcommand{\bbm}{\begin{bmatrix}}
\newcommand{\ebm}{\end{bmatrix}}


%% file: macros/math_defns.tex
\newcommand{\Path}[1]{\xi_{#1}}
\newcommand{\cost}[0]{J}
\newcommand{\costFn}[1]{\cost \left( #1 \right)}

\newcommand{\costSmooth}[0]{\cost_\mathrm{smooth}}
\newcommand{\costFnSmooth}[1]{\costSmooth \left( #1 \right)}

\newcommand{\costOcc}[0]{\cost_\mathrm{occ}}
\newcommand{\costFnOcc}[1]{\costOcc \left( #1 \right)}

\newcommand{\costObs}[0]{\cost_\mathrm{obs}}
\newcommand{\costFnObs}[1]{\costObs \left( #1 \right)}

\newcommand{\costDiv}[0]{\cost_\mathrm{div}}
\newcommand{\costFnDiv}[1]{\costDiv \left( #1 \right)}

\newcommand{\costVis}[0]{\cost_\mathrm{vis}}
\newcommand{\costFnVis}[1]{\costVis \left( #1 \right)}

\newcommand{\costCine}[0]{\cost_\mathrm{cine}}
\newcommand{\costFnCine}[1]{\costCine \left( #1 \right)}

\newcommand{\map}[0]{\mathcal{M}}
\newcommand{\grid}[0]{\mathcal{G}}

%% file: inputs/0_abstract.tex

\begin{abstract}
Aerial cinematography is significantly expanding the capabilities of 
film-makers.
Recent progress in autonomous unmanned aerial vehicles (UAVs) has further increased the potential impact of aerial cameras, with systems that can safely track actors in unstructured cluttered environments.
Professional productions, however, require the use of multiple cameras simultaneously to record different viewpoints of the same scene, which are edited into the final footage either in real time or in post-production.
Such extreme motion coordination is particularly hard for unscripted action scenes, which are a common use case of aerial cameras.
In this work we develop a real-time multi-UAV coordination system that is capable of recording dynamic targets while maximizing shot diversity and avoiding collisions and mutual visibility between cameras. 
We validate our approach in multiple cluttered environments of a photo-realistic simulator, and deploy the system using two UAVs in real-world experiments. 
We show that our coordination scheme has low computational cost and takes only 1.17 ms on average to plan for a team of 3 UAVs over a 10 s time horizon.
Supplementary video: \url{https://youtu.be/m2R3anv2ADE}






\end{abstract}

%% file: inputs/1_intro.tex

\vspace{0mm}
\section{Introduction}
\vspace{0mm}

Flying cameras are revolutionizing industries that require live and dynamic camera viewpoints such as entertainment, sports, and security. 
The use of unmanned aerial vehicles (UAVs) has several advantages in comparison with traditional techniques such as dollies and cranes in terms of cost, efficiency and safety \cite{book1, santamarina2018introduction}. 
Furthermore, recent woks both in industry \cite{mavic2019,skydio2018} and academia \cite{bonatti2020autonomous,jeon2020detection,huang2018act} now allow UAVs to track targets autonomously in cluttered environments.

High-quality cinematographic productions, however, require the composition of multiple viewpoints to form the final footage \cite{bowen2013grammar,arijon1976grammar}, which can be edited \textit{on-the-fly} in the case of live sports events, or more commonly in post-production.
Camera arrangement is a complex problem that requires the production crew to reason about what is going to happen in the scene in terms of actor movements, increase viewpoint diversity, obey cinematographic guidelines (\textit{e.g.} the $180^\circ$ rule, staying on a single side of the scene), and avoid mutual visibility between cameras. 
The problem becomes even more challenging in the case of unscripted scenes in sports or journalistic coverage, where cameras need to be re-arranged more frequently while being subjected to velocity and acceleration constraints, and the event cannot be re-enacted in the case of mistakes.

Despite the large recent progress in single-UAV filming solutions, multi-UAV systems are still scarce.
Existing approaches focus mainly on coordinating vehicles to follow predefined shot types \cite{multi_drone_cine} or pre-planned filming missions \cite{events,auto_planner}, restricting the use of this technology to scripted applications.
We also find work on maximizing actor coverage \cite{coverage}, partially addressing the shot diversity objective, but this approach is restricted to 2D reasoning and static targets. 
None of the existing systems can autonomously coordinate multiple cameras in dynamic and unscripted applications.

As seen in Figure \ref{fig:main}, our work aims to tackle the full multi-UAV coordination problem, targeting cinematography in unscripted and dynamic scenarios.
We propose a system that can generate diverse and artistic shots in real-time, empowering operators with full artistic capabilities of aerial cameras. 
Our main contributions are:

\textbf{1) Problem formulation:} We formalize the multi-UAV coordination problem in term of its principal cost functions. We start with a single-UAV formulation \cite{r1} that considers trajectory smoothness, obstacle avoidance and environmental occlusions, and add new cost terms to maximize shot diversity, avoid mutual visibility between cameras, and to respect high-level user-specified cinematography guidelines;

\textbf{2) Camera coordination:} We propose a greedy framework for multi-UAV coordination that can run in real-time in dynamic scenes. First, a fast centralized planner computes a set of desired viewpoints for each camera, and a decentralized planning system computes the final trajectory for each UAV based on an occupancy map of the environment and the predicted actor trajectory;

\textbf{3) Experimental validation:} We validate our approach in multiple environments using a photo-realistic simulator, and deploy the system in real-world experiments with two UAVs.

%% file: inputs/2_related_work.tex
\section{Related Work} 
\label{sec:related_work}

\textbf{Single-Drone Cinematography:}
There is vast body of academic literature and consumer products on the topic of single-drone cinematography. 
For instance, products such as the DJI Mavic \cite{mavic} and Skydio R1 \cite{skydio2018} can detect and track targets autonomously.
In addition, works such as \cite{r3, auto_1, auto_2, auto_3, auto_4, auto_5, feasible_traj} can follow user-specific artistic guidelines during motion.
We also find works that try to automate the artistic decision-making guidelines as well. For instance, \cite{RL_principles, r1} use deep reinforcement learning to train a deep policy to automatically select the best shot types for a given scene, conditioned on the current actor actions and obstacle field.
Also, \cite{huang2018act} uses the actor's skeleton configuration to automatically position the camera, maximizing the body's projection on the image.

    

\textbf{Multi-Robot Systems:}
There is a rich selection of work on multi-robot systems, ranging from safety and controls~\cite{luo2016distributed,luo2019multi}, planning~\cite{karnad2012modeling,jiang2019multi,liu2016mdp}, target localization~\cite{engin2020active,bayram2017tracking,charrow2014approximate}, exploration~\cite{corah2019distributed,corah2017efficient}, robot swarm task planning \cite{chandarana2018decentralized},  and even theatrical performances~\cite{cappo2018robust,cappo2018online}.
We also find important theoretical work on multi-sensor coordination using efficient greedy methods \cite{krause2008near,roberts2017submodular} that enjoy bounds on sub-optimality when compared to the full exploration of the search space.

\textbf{Multi-Drone Cinematography:}
In the field of UAV cinematography we find a few pioneer works that employ multiple vehicles.  
For instance, \cite{multi_drone_cine} proposes an optimization-based algorithm to coordinate multiple UAVs while accounting for inter-drone collisions and mutual visibility, and requires user-defined paths as guidelines for the motion of each drone.
The work of \cite{auto_planner} takes on additional constraints into a greedy optimization, and maximizes target visibility over time using a team of drones subject to limited battery life.
\cite{coverage} simplifies the actor coverage problem to a 2D space, maximizing target visibility. 
Also focusing on multi-UAV coverage, \cite{3Dreconstruction} optimizes multiple flying cameras trajectories for efficient 3D reconstruction.
In the context of filming outdoor events with dynamic targets, \cite{events} provides an overview of cinematography principles that can be used for filming with multiple drones.


%% file: inputs/3_problem_formulation.tex

\section{Problem formulation} 
\label{sec:problem_formulation}

Our overall task is to optimally control a team of UAVs to film an actor who is moving through an environment with obstacles. 
Similarly to \cite{r1}, we formulate a trajectory optimization problem using cost functions that measure environmental occlusion of the actor, jerkiness of motion and safety. In addition, we introduce new cost components for maximizing shot diversity between UAVs, avoid inter-vehicle visibility, and to allow user-specified cinematographic guidelines.

Let $\Path{qi} : [0,t_f] \rightarrow \real^3  \times SO(2)$ be the trajectory of the \textit{i}-th UAV, i.e., $\Path{qi}(t) = \{x(t), y(t), z(t), \psi_{q}(t)\}$, and $\Xi = \{\Path{q1}, ..., \Path{qn} \}$ be the set of trajectories from $n$ UAVs.
Let $\Path{a} : [0,t_f] \rightarrow \real^3  \times SO(2)$ be the trajectory of the actor, $\Path{a}(t) = \{x(t), y(t), z(t), \psi_{a}(t)\}$, which is inferred using the onboard cameras.
Let grid $\grid: \real^3 \rightarrow [0, 1]$ be a voxel occupancy grid that maps every point in space to a probability of occupancy. Let $\map : \real^3 \rightarrow \real$ be the signed distance values of a point to the nearest obstacle. 
Our mathematical objective is to minimize a cost function $\costFn{\Xi}$ that tends to the following objectives:

\textit{1) Smoothness:} Penalizes jerky motions that may lead to camera blur and unstable flight. Calculated as the sum of costs from individual trajectories:
$\costFnSmooth{\Xi} = \sum_i \costFnSmooth{\Path{qi}}$\\
\textit{2) Occlusion:} Penalizes occlusion of the actor by obstacles in the environment for each camera:
$\costFnOcc{\Xi} = \sum_i \costFnOcc{\Path{qi},\Path{a},\map}$\\
\textit{3) Safety:} Penalizes proximity to obstacles that are unsafe for each UAV:
$\costFnObs{\Xi} = \sum_i \costFnObs{\Path{qi},\map}$\\
\textit{4) Diversity:} Penalizes viewpoints similarities between UAVs. Calculated over the entire set of trajectories:
$\costFnDiv{\Xi,\Path{a}}$\\
\textit{5) Inter-visibility:} Avoids visibility between UAVs:
$\costFnVis{\Xi}$\\
\textit{6) Cinematography guidelines:} Penalizes user-specified undesired viewpoints (\textit{e.g.} high tilt angles):
$\costFnCine{\Xi,\Path{a}}$

We then compose the overall cost function as a linear combination between each component, with relative weights $\lambda$.
The solution $\Xi^*$ is then tracked by each UAV:
\vspace{-2mm}

\begin{equation}
\begin{aligned}
\label{eq:main_cost}
\costFn{\Xi} &=  \begin{bmatrix}
       1 & \lambda_1 & \lambda_2 & \lambda_3 & \lambda_4 & \lambda_5
     \end{bmatrix} 
\begin{bmatrix}
       \costFnSmooth{\Xi} \\
       \costFnOcc{\Xi} \\
       \costFnObs{\Xi} \\
       \costFnDiv{\Xi}\\
       \costFnVis{\Xi}\\
       \costFnCine{\Xi}
     \end{bmatrix} \\
\Xi^* &= \argminprob{} \quad \costFn{\Xi}
\end{aligned}
\end{equation}
\vspace{-2mm}


%% file: inputs/3_approach.tex

\section{Multi-Camera Coordination} 
\label{sec:approach}

We now detail the methods we use for camera coordination in the multi-UAV system.
As displayed in Equation~\ref{eq:main_cost}, our overall objective function involves the minimization of $6$ sub-objectives, which often conflict with one another. 
Our goal is to formulate an algorithm that works in real-time, in unscripted scenes, and that can deal with a state space which grows exponentially in complexity with the number of UAVs. 
Given this challenge, we prefer to find fast solutions which are only locally optimal as opposed to globally optimal trajectories that take a long time to compute.

\begin{figure}[htp]
    \centering
    \includegraphics[width=0.48\textwidth]{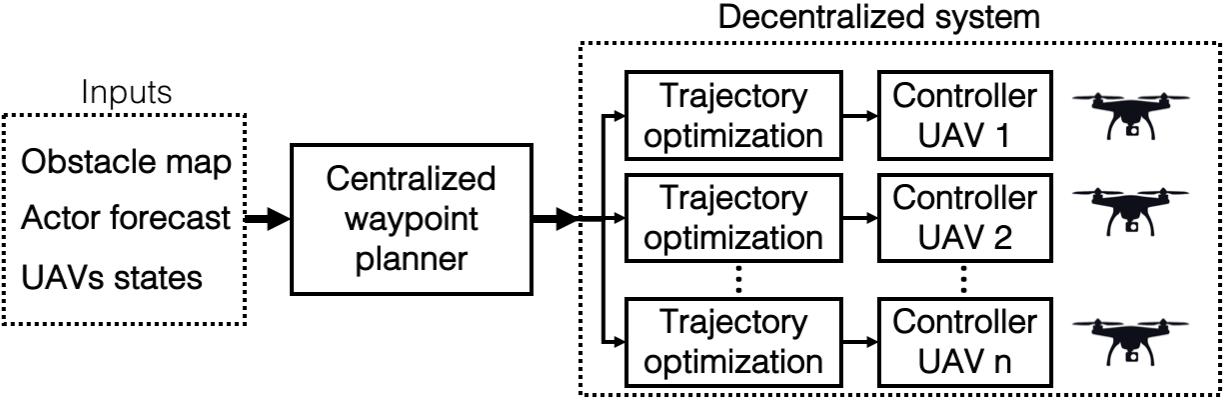}
    \caption{\small System overview: a centralized waypoint planner computes discrete camera positions. Next, a decentralized planning system optimizes smooth trajectories for each UAV.
    \vspace{-3mm}}
    \label{fig:system}
\end{figure}

To address the time complexity issue, we break down our method into three main subsystems that operate together.
First, a centralized motion planner (Sec.~\ref{subsec:centralized}) coordinates desired positions for all cameras simultaneously.
Next, a decentralized motion planner network (Sec.~\ref{subsec:decentralized}) computes the final trajectories for each specific UAV.
Finally, an image selection module (Sec.~\ref{subsec:img_selection}) chooses the best live image to be displayed out of all cameras. 
Alternatively, the final image selection can be manually performed in post-production. 
Fig.~\ref{fig:system} depicts the system diagram.

\subsection{Centralized Multi-Camera Planning}
\label{subsec:centralized}

Our centralized planning system parametrizes trajectories as waypoints, i.e., $\Path{} \in \real^{T \times 3}$, where $T$ is the number of time steps. 
We assume that the heading dimension $\psi(t)$ is set to always point the drone from $\Path{qi}(t)$ towards the actor in $\Path{a}(t)$, which can be achieved independently of the aircraft's translation by rotating the UAV's body and camera gimbal.
We parametrize the state-space of all possible camera positions using spherical coordinates $\{\rho, \theta, \phi\}$ centered on the actor's location: 
\vspace{-4mm}

\begin{equation}
	\begin{aligned}
		&\Path{qi}(t) = \Path{a}(t) + \rho 
		\begin{bmatrix}
		 cos(\psi_a + \theta) sin(\phi)\\
		 sin(\psi_a + \theta) cos(\phi)\\
		 cos(\phi)
		\end{bmatrix}\\
	\end{aligned}
\end{equation}


\textbf{Space discretization: } 
We define a discrete state-space lattice $S$ that contains all possible camera positions distributed as $|S| = 576$ points in a half-sphere above ground.
\begin{wrapfigure}{r}{0.45\linewidth}
    \centering
    \includegraphics[width=1.0\linewidth]{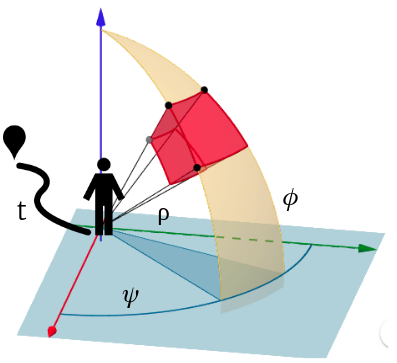}
    \caption{\small Discrete spherical state space centered on actor.
    \vspace{-2mm}}
    \label{fig:state_space}
\end{wrapfigure}
Based on cinematographic guidelines \cite{arijon1976grammar,bowen2013grammar} we equally divide the yaw coordinates $\theta$ into $16$ values between $[0, 2\pi]$, the tilt angles $\phi$ into $6$ values between $[0, \pi/2]$, and the distance to actor within the set $\rho \subseteq \{2, 3, 4, 5, 6, 7\}$, ranging from close-up to long shots. 
In addition, we discretize the trajectory's time into $5$ steps equally spaced every $2$ seconds, forming a $10$-second planning time horizon.

\textbf{Cost functions: }
Next we mathematically formulate the cost functions that optimized by the centralized camera planner for the set of UAV trajectories $\Xi$: 


\textit{i) Shot diversity cost: }
The shot diversity cost prevents the planner to allocate the $n$ camera positions close to each other up until a maximum distance $d_{max}^{div}$:
\vspace{-0mm}
\begin{equation}
\label{eq:cost_div}
\begin{aligned}
    &\costFnDiv{\Xi} = \sum_{\tau=1}^{T} \sum_{i=1}^{n} \sum_{j=1}^{n} D({\Path{qi}(\tau), \Path{qj}(\tau)}),\\
    \text{where} \quad & D(p_1,p_2) = 
    	\begin{cases} 
    		1 & \mbox{if } d < d_{min}^{div} \\ 
            \frac{d}{d_{max}^{div} - d_{min}^{div}} & \mbox{if } d_{min}^{div} < d < d_{max}^{div}\\
            0 & \mbox{if } d > d_{max}^{div}
        \end{cases}
\end{aligned}
\end{equation}




\textit{ii) Inter-drone collision cost: }
Analogously to $\costDiv$ in Eq.~\ref{eq:cost_div}, we define the inter-drone collision cost as a measure of distance between UAVs. 
The user can, however, choose a different cost weight $\lambda$ and safety distance $d_{max}^{col}$ to penalize collisions.


\textit{iii) Inter-drone visibility:}
We express the inter-drone visibility cost as a binary measure $V(p_i,p_j) \subseteq \{0,1\}$ of whether point $j$ is visible from position $i$.
We model the visible area as a cone in space using the camera's diagonal field-of-view angle, and position its center line towards the actor's position:
\vspace{-2mm}
\begin{equation}
\label{eq:cost_vis}
    \costFnVis{\Xi} = \sum_{\tau=1}^{T} \sum_{i=1}^{n} \sum_{j=1}^{n} V({\Path{qi}(\tau), \Path{qj}(\tau)})
    \vspace{-2mm}
\end{equation}

We pre-compute all mutual visibility values into a look-up table to reduce planning times, which is possible in our discrete state-space model.





\begin{figure}[ht]
     \centering
     \begin{subfigure}[b]{0.47\linewidth}
         \centering
         \includegraphics[width=\linewidth]{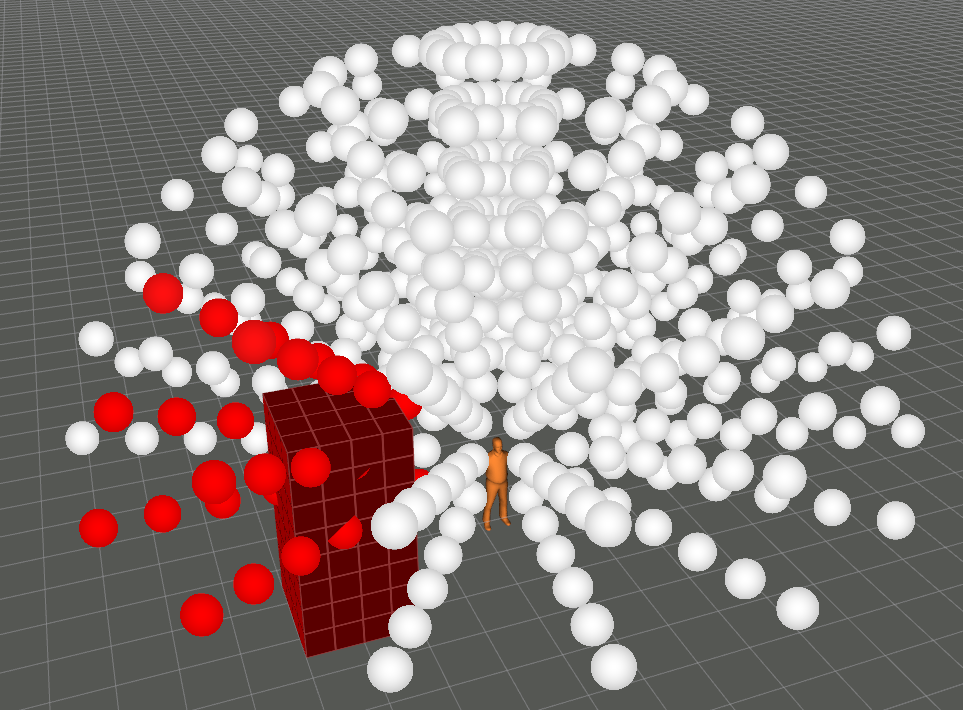}
         \caption{}
         \label{fig:grid_sphere}
     \end{subfigure}
     \hfill
     \begin{subfigure}[b]{0.51\linewidth}
         \centering
         \includegraphics[width=\linewidth]{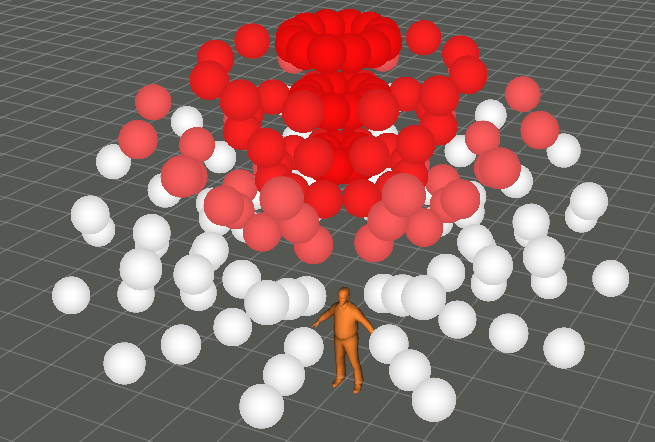}
         \caption{}
         \label{fig:prior_overhead}
     \end{subfigure}
        \caption{\small
        (a) Visualization of occupancy and occlusion avoidance costs in the cells of the spherical grid $\grid_{s}^{t}$.
        (b) Example of a cinematography guideline prior that penalizes overhead shots. Red  spheres  represent  regions of high cost.
        \vspace{-5mm}}
        \label{fig:three graphs}
\end{figure}

\textit{iv) Obstacle and occlusion avoidance: }
In order to keep all UAVs safe and to maintain visibility of the actor at all times, we must reason about the role of obstacles in the environment.
First, we transform the environment's occupancy grid into a time-dependent spherical domain centered around the actor $\grid \rightarrow \grid_{s}^{t} \in [0,1]$, as shown in Fig.~\ref{fig:grid_sphere}. We then compute the obstacle avoidance cost by summing the occupancy likelihood of all cells within a radius $r_{max}$ of each UAV. We also calculate the occlusion cost as a measure of occupancy along a line $l_i(\tau) = \tau\Path{qi}(t) + (1-\tau)\Path{a}(t)$ between UAV and actor:
\vspace{-5mm}

\begin{equation}
\begin{aligned}
\label{eq:cost_obs_and_occ}
    \costFnObs{\Xi} &= \sum_{\tau=1}^{T} \sum_{i=1}^{n} \int_0^{r_{max}} \grid_{s}^{\tau}({\Path{qi}(\tau)}) \ d(\text{volume})\\
    \costFnOcc{\Xi} &= \sum_{\tau=1}^{T} \sum_{i=1}^{n} \int_0^{1}\grid_{s}^{\tau}({l_i(\tau)}) \ d\tau
    \vspace{-2mm}
\end{aligned}
\end{equation}









\textit{v) Cinematography guidelines as cost prior: }
We also allow operators to manually specify a cost prior $\costCine$ for each cell, in case they wish to follow specific cinematographic guidelines. 
For example, Fig.~\ref{fig:prior_overhead} shows an example where we define overhead shots as undesired.

\textbf{Greedy optimization: }
The centralized planning problem for multi-camera optimization over multiple time steps proves to be NP-hard, similarly to other optimal sensor placement works \cite{greedy1,krause2006near,arora2017randomized}.
In order to make the computation tractable in real-time and avoid a combinatorial problem, we develop a greedy optimization approach that computes the optimal trajectory 
$\Path{q\,i}^{*\text{greedy}}$
for each UAV sequentially, fixing all previously calculated trajectories 
$\{ \Path{q\,1}^{*\text{greedy}}$, ..., $\Path{q\,i-1}^{*\text{greedy}} \}$. 

\begin{wrapfigure}{r}{0.45\linewidth}
    \centering
    \includegraphics[width=1\linewidth]{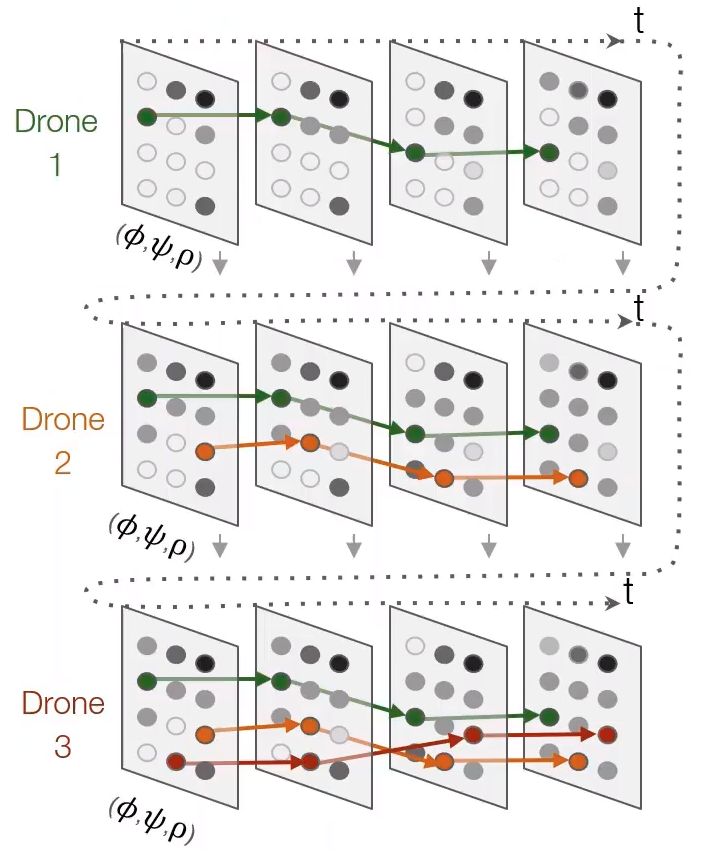}
    \caption{\small Greedy planner.
    \vspace{-4mm}}
    \label{fig:greedy}
\end{wrapfigure}

At each stage $i$ we employ a backward induction dynamic programming algorithm \cite{sutton2018reinforcement,bertsekas1995dynamic,lavalle2006planning} to find the optimal cost-to-go for of each state $s \subseteq S$ at all time steps, analogously to a value iteration algorithm.
To do so, we build a cost map $C: S \rightarrow \real^{|S|}$ that contains the cost of all states, and a cost-to-go map $V: S \rightarrow \real^{|S|}$.
In order to make transitions between cells dynamically feasible for the real vehicle, we only allow expansions to neighboring cells in the spherical grid.
Since we operate in a discrete state-space with a small branching factor and deterministic transitions, a single backwards pass quickly yields the optimal solution.
Finally, we build the full trajectory $\Path{q\,i}^{*\text{greedy}}$ by selecting neighboring cells with the least cost-to-go at consecutive time steps, starting at the UAV's initial position $S_i^0$.
We update the cost manifold after each drone is added, since inter-visibility, shot diversity, and collision costs are recalculated. 
Algorithm~\ref{alg:greedy_alg} details the process:

\small
\begin{algorithm}[ht]
\caption{\small Compute greedy traj. set $\Xi = \{\Path{q1}, ..., \Path{qn} \}$}
\label{alg:greedy_alg}
\SetAlgoLined
 Initialize $\Xi_{\textnormal{greedy}} = \{\}$ \Comment{\footnotesize empty set}
 \For{\textnormal{\textbf{each}} \textnormal{UAV} $i$}{
  $C$ $\gets$ $J(S, \Xi_{\textnormal{greedy}})$ \Comment{\footnotesize update entire cost map}
  $V$ $\gets$ Backward\_Induction($C$) \Comment{\scriptsize update cost-to-go}
  $\Path{q\,i}^{*\text{greedy}}$ $\gets$ Optimal\_Path($V$, $S_i^0$)\;
  Append $\Path{q\,i}^{*\text{greedy}}$ to $\Xi_{\textnormal{greedy}}$\;
 }
 \textbf{return} $\Xi_{\textnormal{greedy}}$
\end{algorithm}
\vspace{-3mm}
\normalsize

\subsection{Decentralized UAV Trajectory Optimization and Tracking}
\label{subsec:decentralized}

After calculating the greedy-optimal set of UAV trajectories $\Xi_{\textnormal{greedy}}$ using the centralized planner, we post-process each output using a decentralized planning system.
Our objective here is to obtain smoother individual trajectories at a finer time discretization. 
While the original waypoints were spaced every $2$ seconds over a $10$-second horizon, here we achieve finer resolutions with $0.5$ s granularity in local planning, and $0.02$ s for trajectory tracking.
We employ the single-UAV local planner described in \cite{r1}, which uses covariant gradient descent to produce locally optimal trajectories while considering the costs of smoothness, obstacle and occlusions avoidance, and a desired artistic trajectory, which we define as $\Path{q\,i}^{*\text{greedy}}$. 
In addition, each local planner receives the expected waypoints of all remaining vehicles, and avoids positioning its trajectory within 1m of other UAVs.
We run the local planner at $5$ Hz, and use a PID controller at $50$ Hz.
\vspace{-0mm}

\subsection{Live Image Selection}
\label{subsec:img_selection}
\vspace{-0mm}
Even though our system produces multiple image streams, most filming applications display a single camera to the viewer at a time.
The final movie can be produced with manual post-processing, or in the case of live streams it requires a human to make selections while viewing all videos simultaneously.
Here, we propose a method for automatic selection of live images based on a subset of our cost functions. 
We compute a image quality score $Q_i$ for each camera stream, calculated as a weighted sum of the inter-drone visibility cost $\costVis$ and of the prior cinematographic guidelines $\costCine$. 
We select the current camera with the highest score, and once a camera is chosen we exponentially decay its score to foster viewpoint diversity, and gradually return it to the original value once another image is chosen.
Based on cinematography literature we set minimum and maximum limits for shot lengths of $3$ and $8$ seconds respectively, which are reasonable units of length for individual action shots \cite{mahadani_mahadanii_2015,cutting2010attention}.

%% file: inputs/4_exp.tex
\section{Experimental Results} 
\label{sec:results}
Here we detail the simulated and real-world experiments that validate our multi-UAV cinematography system. 
Additional visualizations are shown in the supplementary video.

\subsection{Simulation experiments}
\label{subsec:exp_sim}

\textbf{Experimental setup: }
We record all simulated data using a drone in a photo-realistic environment,
AirSim \cite{shah2018airsim}, coupled with a custom ROS interface \cite{ros}. 
As seen in the supplementary video, our simulated scenes consists of
an animated character walking around a suburban environment, surrounded by obstacles such as trees, buildings, and posts.

\begin{figure*}[]
     \centering
     \begin{subfigure}[b]{0.325\textwidth}
         \centering
         \includegraphics[width=\textwidth]{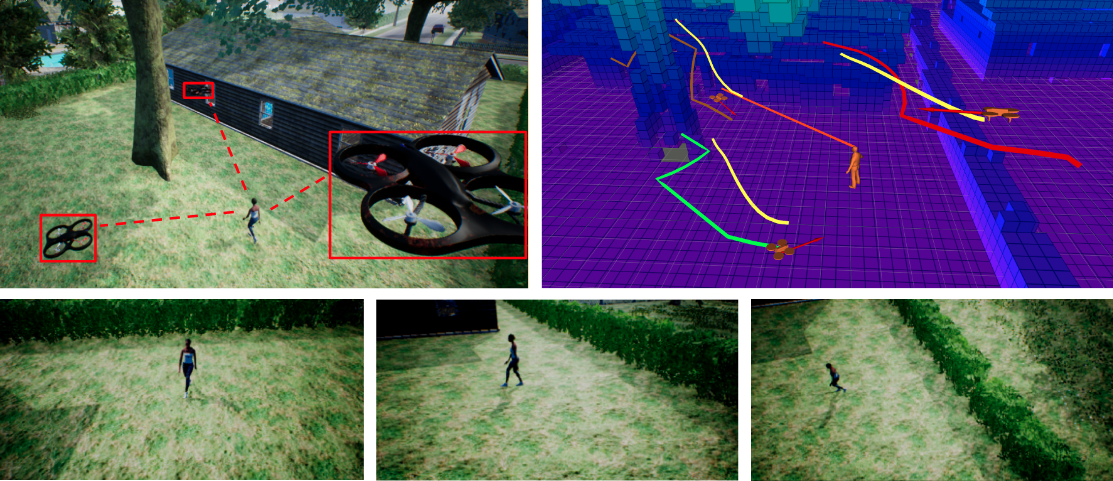}
         \caption{}
         \label{fig:env_rural}
     \end{subfigure}
     \hfill
     \begin{subfigure}[b]{0.325\textwidth}
         \centering
         \includegraphics[width=\textwidth]{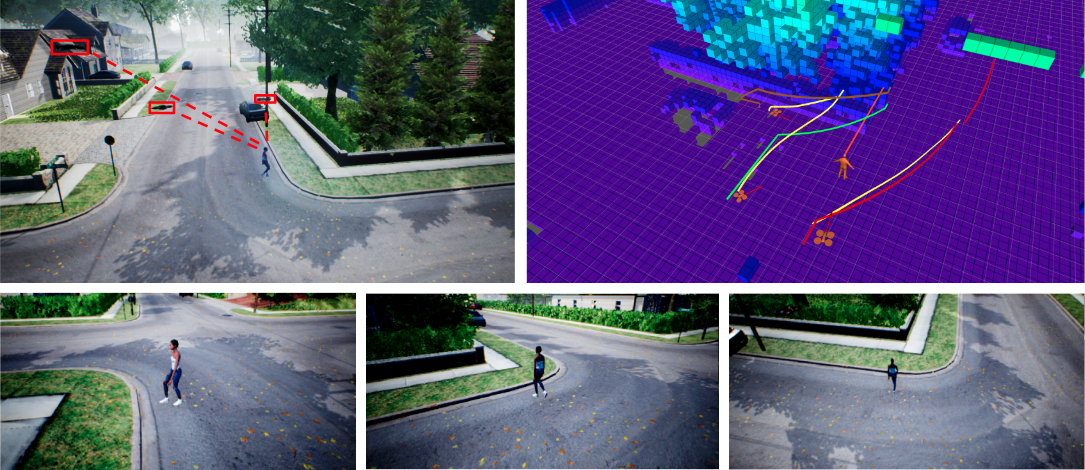}
         \caption{}
         \label{fig:env_city}
     \end{subfigure}
     \hfill
     \begin{subfigure}[b]{0.325\textwidth}
         \centering
         \includegraphics[width=\textwidth]{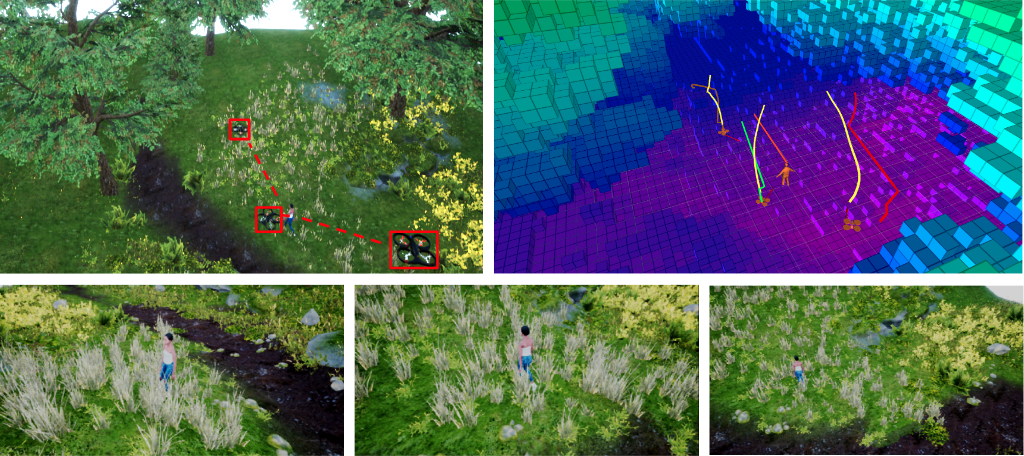}
         \caption{}
         \label{fig:env_forest}
     \end{subfigure}
        \caption{\small Diverse range of simulated environments where we tested our multi-UAV system: a) narrow gaps, b) suburban environment with sparse tall obstacles, and c) dense foliage.
        \vspace{-4mm}}
        \label{fig:three_envs}
\end{figure*}

\textbf{E1) Viewpoint diversity validation: }
This first experiment's objective was to validate the assumption of the relation between shot diversity and artistic quality by quantifying the benefits of the employing multiple UAVs for aerial cinematography as opposed to the use of a single camera.
To do so, we generated a set of $5$ videos with $12$ seconds of length of an actor walking around a park.
The first clip featured a single-camera back shot for its entire duration, while all other clips alternated between the back shot and a secondary viewpoint every $3$ seconds. 
We chose the secondary viewpoints to be either $45^{\circ}$, $90^{\circ}$, $135^{\circ}$, or $180^{\circ}$ for each of the other four videos respectively. 

For this survey we recruited $15$ participants using Amazon Mechanical Turk (MTurk) \cite{turkamazon}, who were compensated for their time.
After being approved on a short qualifying task, each participant viewed a total of 12 pairs of videos, one being the constant back shot clip, and the other being one of the multi-viewpoint clips. 
Videos were played synchronously three times, and after watching at least once participants answered: ``\textit{Which video is more enjoyable?}''
Each clip was compared 15 times against the baseline single-camera shot.

\begin{figure}[ht]
    \centering
    \includegraphics[width=1.0\linewidth]{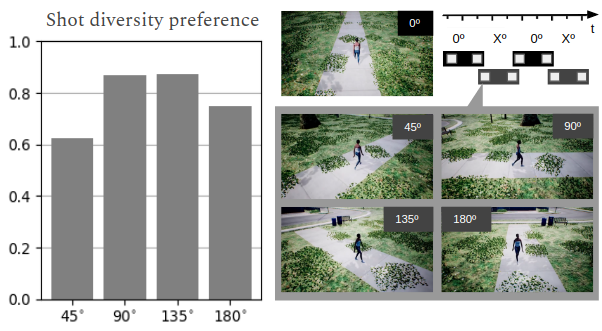}
    \caption{\small User study to evaluate viewer preference for videos with diverse viewpoints. We ask users to compare how enjoyable a single back shot view is against clips with additional viewpoints. Vertical bar shows percentage of users that prefer the video with a secondary viewpoint, where each bar corresponds to an angle value.
    \vspace{-2mm}}
    \label{fig:shot_diversity}
\end{figure}

Figure \ref{fig:shot_diversity} displays the survey results, where the height of each bar represents the percentage of users that rated the multi-UAV clip as being more enjoyable than the single-UAV clip.
As expected, we found that an overwhelming majority of users prefer to watch a scene with viewpoint diversity.

\textbf{E2) Robustness across environments: }
We tested our approach in a diverse set of photo-realistic simulations, ranging from open to cluttered environments, and with groups of UAVs containing 2 to 5 vehicles. 
As shown in Figure \ref{fig:three_envs}, our approach is able to successfully coordinate the teams of UAVs in a wide spectrum of scenarios.  
For instance, in \ref{fig:env_rural} the actor passes trough a narrow gap (2m) that the UAVs need to avoid, and in \ref{fig:env_city} she walks next to a extensive line of trees (1m apart), which forces the UAVs to move to the right side of the actor. In addition, in \ref{fig:env_forest} we see the UAVs navigating within dense and unstructured foliage tunnels.
 
The tests indicate that our system is able to adapt to a diverse range of environments and actor motions. Furthermore, we verify that the final clips follow our objective functions and present diverse viewpoints with little inter-drone visibility.

\textbf{E3) Scaling performance: }
We conducted a set of experiments to quantify our centralized planner's performance as it scales to larger teams of UAVs and larger state-spaces. 
This analysis is important because keeping low computational costs is vital for real-time performance.

First, we measured the planning time and computer resources consumption for finer discretizations of the state space (Table \ref{table:performance}).
We note that memory usage grows proportionally to the square of the number of cells in the state space because we employ a precomputed lookup table for the inter-drone visibility and shot diversity costs. 
However, our implementation offers large benefits in terms of computing time and CPU usage, which grow linearly.

\begin{table}[ht]
\centering
\begin{tabular}{ll|lll}
\begin{tabular}[c]{@{}l@{}}State space\\ ($n_{\psi}$,$n_{\phi}$,$n_{\rho}$)\end{tabular} &
  \begin{tabular}[c]{@{}l@{}}Computed\\ states\end{tabular} &
  \begin{tabular}[c]{@{}l@{}}Planning time\\ for 3 drones [ms]\end{tabular} &
  \begin{tabular}[c]{@{}l@{}}CPU[\%]\\ (1 core)\end{tabular} &
  \begin{tabular}[c]{@{}l@{}}Memory\\ use [MB] \end{tabular} \\ \cline{1-5}
(3,3,8)    & {\color[HTML]{000000} 360} & 0.17+-0.03   & 22  & 35    \\
(16,6,6)   & 2880                       & 1.17+-0.18   & 25  & 37.5  \\
(24,9,9)   & 9720                       & 4.65+-0.47   & 28  & 64    \\
(32,12,12) & 23040                      & 16.70+-1.26  & 34  & 197   \\
(40,15,15) & 45000                      & 19.56+-0.83  & 38  & 655   \\
(48,18,18) & 77760                      & 51.71+-2.04  & 52  & 1840  \\
(52,21,21) & 114660                     & 116.28+-6.98 & 75  & 4693  \\
(64,24,24) & 184320                     & 228.01+-8.85 & 100 & 10150
\end{tabular}
\caption{\small Performance of the greedy planner for different discretizations of the State space. We use 3 UAVs with a 5 time-steps horizon, and re-plan at 5 Hz. \label{table:performance}}
\end{table}

The discretization we use for most of our experiments ($n_{\psi}$=16,$n_{\phi}$=6 and $n_{\rho}$=6) presents a great trade-off, with a reasonable space discretization and a low computation time of 1.17ms, while only consuming 37.5 MB of RAM.
As a comparison, we implemented an optimal brute-force planner and tested it using the same spatial discretization and number of drones. 
Using only 2 time steps, the non-greedy planner took 16.4 seconds to find the optimal solution. The planning time jumps to 2h:30min when we consider 3 time steps. These results show the importance of adopting a greedy strategy to solve a NP-hard problem in real-time.

\begin{figure}[ht]
    \centering
    \includegraphics[width=1.0\linewidth]{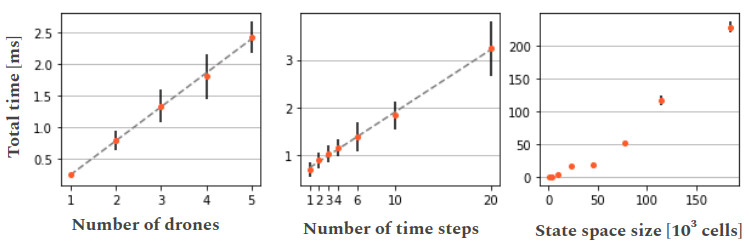}
    \caption{\small Time required by the high-level planner to compute the trajectories of the drones: (a) given the number of vehicles, (b) given the number of time steps computed an (c) given the number of states in the artistic domain ($n_{\psi} \cdot n_{\phi} \cdot n_{\rho} \cdot n_t$)
    \vspace{-3mm}}
    \label{fig:time_charts}
\end{figure}

We also evaluated how the planning time increase with a larger number of drones and time steps. 
As expected, our greedy solution has a linear growth of complexity (Fig.~\ref{fig:time_charts}).

\subsection{Real-world results}
\label{subsec:exp_real}

\textbf{Experimental setup: }
For the real-world experiments we used two unmodified Parrot Bebop 2 UAVs with an integrated electronic gimbal, a WiFi router and a standard desktop PC (Intel i7-8700 CPU and Nvidia GTX1060 GPU).
All communication between the drones and PC were handled via a ROS interface with the Bebop SKD. 
We transmitted images from the UAVs to the base station for the actor's visual detection and trajectory forecasting, and sent velocity commands back to the UAV after planning. 

\begin{figure}[ht]
    \centering
    \includegraphics[width=1.0\linewidth]{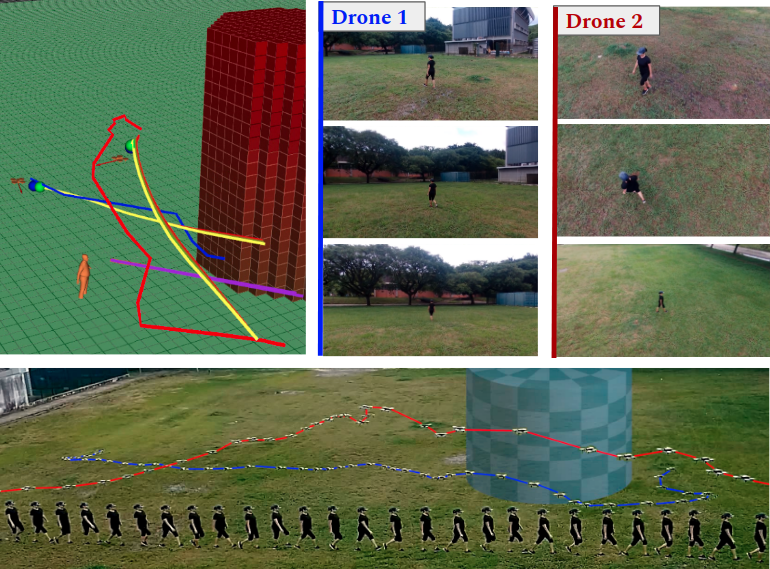}
    \caption{\small Real-life flight among a simulated obstacle. On the left, the outputted sequence waypoints for each drone (red and blue) guide the locally optimized paths (yellow) towards a region free from obstacles and free of inter-drone visibility. Diverse shots can be seen during flight.
    \vspace{-2mm}}
    \label{fig:study_case}
\end{figure}

\textbf{E4) Flight among obstacle: } We tested the system by recording a moving actor with 2 drones in an environment with a virtual obstacle. 
Figure \ref{fig:study_case} depicts how the high-level central planner interacts with the local trajectory optimization method to calculate the final UAV paths. 
\textit{Drone 1} is optimized first, and takes a trajectory close to, but sufficiently far from the obstacle. 
\textit{Drone 2} is optimized next by the central planner. When it reaches the vicinity of the virtual obstacle, the planner brings its ideal path to a higher elevation, above the first drone's path, in order to avoid colliding with the obstacle while keeping the actor visible and avoiding visibility of \textit{drone 1}.
The central planner's output for drones 1 and 2 are colored in blue and red respectively.
The UAV's respective local planners receive the discrete high-level path and optimize it, creating a smooth trajectory output.
After \textit{drone 2} reaches the left side of the actor, it switches to a slightly tilted and distant shot due to the shot diversity cost.

\textbf{E5) Dynamic actor: } We evaluated the system's capability to track and record an actor performing dynamic movements with abrupt motion changes, such as a person playing soccer. As seen in the supplementary video, both drones were able to safely keep the actor in frame through the entire test while exploring diverse viewpoints with low inter-drone visibility.

\begin{figure}[ht]
    \centering
    \includegraphics[width=1.0\linewidth]{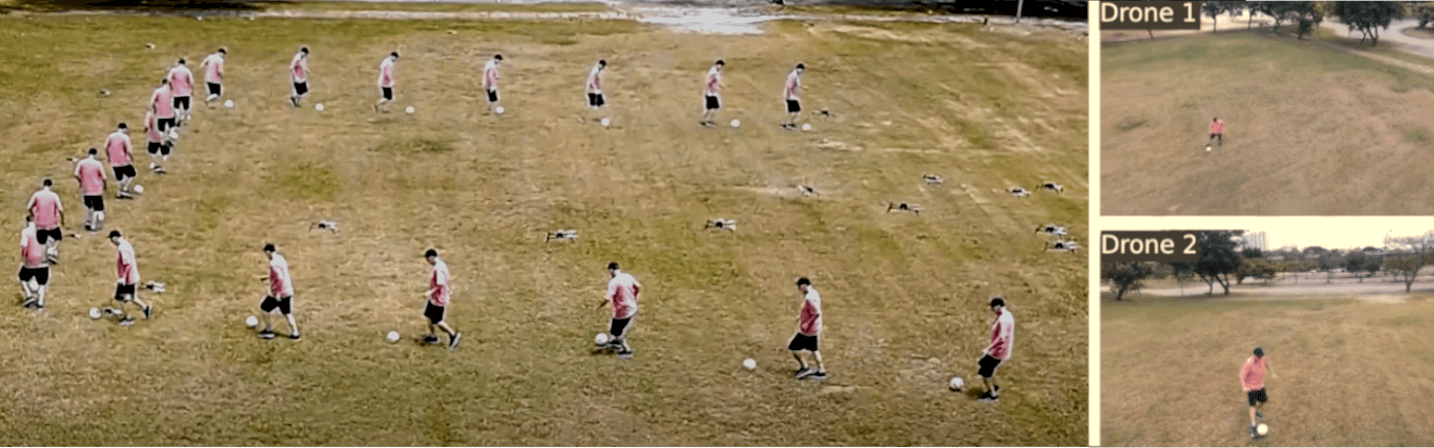}
    \caption{\small Real-life flight following a highly dynamic actor playing soccer. The drones are able to keep up with abrupt motion motion changes while filming the actor.
    \vspace{-4mm}}
    \label{fig:fut}
\end{figure}

%% file: inputs/5_discussion.tex

\section{Conclusion and Discussion}
\label{sec:discussion}

In this paper we present a system for real-time coordination of aerial cameras for autonomous cinematography in dynamic and unscripted scenarios.
First, we formalize the multi-UAV filming problem in terms of its main objectives: maximizing shot diversity, avoiding inter-vehicle visibility and obeying high-level cinematographic guidelines.
Next, we develop a two-step approach for calculating the trajectory of each UAV, based on an efficient centralized greedy planner for viewpoint selection coupled with a decentralized trajectory optimizer to calculate smooth trajectories.
We validate our system in multiple simulated and real-world experiments, and show that it can successfully control a team of UAVs.
Additionally, we provide insights into how our methods can scale for larger numbers of UAVs in terms of planning time and memory. 

We note that the nature of trajectories generated with our system is highly dependent on the combination of relative weights between costs. 
Different applications may require distinct weighing schemes.
For instance, for a journalistic coverage the operator would likely increase the penalty on inter-drone visibility, while this weight could be negligible for footage generated in a 3D motion reconstruction application.

We are interested in extending this research in multiple future directions.
Despite using a decentralized trajectory optimization framework, our experiments in this paper were performed on a single desktop PC, and commands were then transmitted to each UAV individually in real time.
We are currently working on a new multi-UAV system with onboard processors, where one master drone will run the fast centralized planner and then communicate with all other UAVs via a mesh network so that each drone can optimize its own trajectories.
In addition, we plan to employ our system in novel applications such as 3D reconstruction of scenes in the wild. 
Even though multi-view systems are already well developed for indoors environments \cite{joo2015panoptic}, creating such systems for capturing images in natural environments is still an open research field.
